\documentclass{article}

 \usepackage[preprint]{neurips_2025}

\usepackage[utf8]{inputenc} % allow utf-8 input
\usepackage[T1]{fontenc}    % use 8-bit T1 fonts
\usepackage{hyperref}       % hyperlinks
\usepackage{url}            % simple URL typesetting
\usepackage{booktabs}       % professional-quality tables
\usepackage{amsfonts}       % blackboard math symbols
\usepackage{nicefrac}       % compact symbols for 1/2, etc.
\usepackage{microtype}      % microtypography
\usepackage{xcolor}         % colors

\usepackage{tikz}
\usetikzlibrary{bayesnet}

\usepackage{environ}
\makeatletter
\newsavebox{\measure@tikzpicture}
\NewEnviron{scaletikzpicturetowidth}[1]{%
  \def\tikz@width{#1}%
  \begin{lrbox}{\measure@tikzpicture}%
  \BODY
  \end{lrbox}%
  \pgfmathparse{#1/\wd\measure@tikzpicture}%
  \BODY
}
\makeatother

\usepackage{xcolor}
\usepackage{multirow}
\usepackage{placeins} % Required for \FloatBarrier
\usepackage{wrapfig} % For wrapping text around figures
\usepackage{amsmath}
\usepackage{amssymb}
\usepackage{subcaption}

\usepackage{algorithm}
\usepackage{algorithmic}

\usepackage{graphicx}
\usepackage{colortbl}
\usepackage{array}

\newcommand{\bs}{\boldsymbol}
\newcommand{\eg}{{\em e.g.}}
\newcommand{\ie}{{\em i.e.}}

\renewcommand{\k}{{\mathrm{k}}}

\newcommand{\x}{{\bs{x}}}
\renewcommand{\xi}{\x_{i}}
\newcommand{\xj}{\x_{j}}

\newcommand{\z}{{\bs{z}}}
\newcommand{\zi}{\z_{i}}
\newcommand{\zj}{\z_{j}}

\newcommand{\h}{{\bs{h}}}
\newcommand{\hi}{\h_{i}}

\newcommand{\params}{{\bs \theta}}

\newcommand{\M}{\mathcal{M}}
\newcommand{\Mmodel}{\M_{\params}}
\newcommand{\Msamp}{\M_{\mathcal{S}}}

\newcommand{\pjoint}{\mathcal{P}}

\newcommand{\pdec}{p}
\newcommand{\penc}{q}

\newcommand{\Mdec}{\pdec_{\params}}
\newcommand{\Menc}{\penc_{\params}}

\newcommand{\MIM}{{MIM }}

\newcommand{\MIMloss}{\mathcal{L}_\text{MIM}}
\newcommand{\AMIMloss}{\mathcal{L}_\text{A-MIM}}
\newcommand{\EAMIMloss}{\hat{\mathcal{L}}_\text{A-MIM}}

\newcommand{\E}[2]{\mathbb{E}_{#1}\left[#2\right]}

\newcommand{\CE}[2]{CE \left( \, #1 \,,\, #2 \, \right)}

\newcommand{\SIM}[2]{s\left( #1, #2 \right)}
\newcommand{\EXPSIM}[2]{g\left( #1, #2 \right)}

\newcommand{\Pmix}{\frac{1}{2} \left (\Mdec(\k | \z, \x)\, \Mdec(\x | \z)\, \pjoint(\z) + \Menc(\k | \z, \x)\, \Menc(\z | \x)\,
\pjoint(\x) \right )}

\title{Contrastive MIM: A Contrastive Mutual Information Framework for Unified Generative and Discriminative Representation Learning}

\author{%
  Micha Livne \\
  NVIDIA\\
  \texttt{mlivne@nvidia.com} \\
  \textbf{WORKING DRAFT}
}

\begin{document}

\maketitle

% sections
\begin{abstract}

Learning representations that generalize well to unknown downstream tasks is a central challenge in representation learning. Existing approaches such as contrastive learning, self-supervised masking, and denoising auto-encoders address this challenge with varying trade-offs. In this paper, we introduce the \textit{contrastive Mutual Information Machine} (cMIM), a probabilistic framework that augments the Mutual Information Machine (MIM) with a novel contrastive objective. While MIM maximizes mutual information between inputs and latent variables and encourages clustering of latent codes, its representations underperform on discriminative tasks compared to state-of-the-art alternatives. cMIM addresses this limitation by enforcing global discriminative structure while retaining MIM's generative strengths.

We present two main contributions: (1) we propose cMIM, a contrastive extension of MIM that eliminates the need for positive data augmentation and is robust to batch size, unlike InfoNCE-based methods; (2) we introduce \textit{informative embeddings}, a general technique for extracting enriched representations from encoder--decoder models that substantially improve discriminative performance without additional training, and which apply broadly beyond MIM.

Empirical results demonstrate that cMIM consistently outperforms MIM and InfoNCE in classification and regression tasks, while preserving comparable reconstruction quality. These findings suggest that cMIM provides a unified framework for learning representations that are simultaneously effective for discriminative and generative applications.

\end{abstract}

\section{Introduction} \label{sec:intro}

Learning representations that remain effective across unknown downstream tasks is a central challenge in representation learning. Prominent approaches addressing this challenge include contrastive learning (\eg, \cite{chen2020simple,oord2018cpc}),
self-supervised masking (\eg, \cite{devlin2018bert}), and denoising auto-encoders (\eg, \cite{bengio2013generalized}).

In this work, we propose a new method, \textit{cMIM} (Contrastive MIM), designed to learn representations that are broadly useful for downstream applications. cMIM integrates a contrastive learning objective with the Mutual Information Machine (MIM) framework introduced by \cite{livne2019mim}. MIM is a probabilistic auto-encoder that maximizes mutual information between inputs and latent representations while clustering the latent codes. However, preliminary results suggest that MIM alone yields representations less effective for discriminative tasks compared to state-of-the-art alternatives. Our cMIM framework directly addresses this limitation by incorporating contrastive learning.

Our main contributions are as follows:
\begin{enumerate}
\item We propose a contrastive extension to the Mutual Information Machine (MIM) that enables learning discriminative representations without requiring data augmentation (no explicit ``positive'' pairs) and with reduced sensitivity to the number of negative samples (typically determined by batch size).
\item We introduce \textit{informative embeddings}, a generic method for extracting embeddings from encoder--decoder models. This technique improves discriminative downstream performance without additional training and applies broadly to pre-trained encoder--decoder architectures.
\end{enumerate}

By combining generative modeling with contrastive objectives, cMIM provides a unified framework effective for both generative and discriminative tasks. Empirical results show that cMIM produces representations that retain MIM's generative capacity while significantly improving performance in downstream discriminative settings.

\section{Formulation} \label{sec:formulation}

% \begin{figure}[t]
%     \centering
%     \input{diag/encoder-decoder}
%     \vspace*{-0.05cm}
%     \caption{A \MIM model learns two factorizations of a joint distribution:
%     (a) encoding; (b) decoding factorizations; and (c) the estimated joint distribution 
%     (an undirected graphical model).
%     }
%     \label{fig:mim-model}
% \end{figure}

% \begin{figure}[t]
%     \centering
%     \input{diag/x-z-k}
%     \vspace*{-0.05cm}
%     \caption{We extend the \MIM model with additional binary variable $\k$, and present the two factorizations of a joint distribution:
%     (a) encoding; (b) decoding factorizations.
%     }
%     \label{fig:cmim-model}
% \end{figure}

\begin{figure}[t]
    \centering

    \begin{subfigure}[t]{0.48\textwidth}
        \centering
        \begin{tikzpicture}

    % Define nodes
    
    \node[latent]                (zenc) {$\z$};
    \node[obs, below=of zenc]                (xenc) {$\x$};
    
    \node[latent, right=of zenc]               (zdec) {$\z$};
    \node[obs, below=of zdec]                (xdec) {$\x$};
    
    \node[latent, right=of zdec]             (z) {$\z$};
    \node[obs, below=of z]                   (x) {$\x$};
    
     \node[const, below=of xenc, yshift=0.65cm]  {(a)} ; %
     \node[const, below=of xdec, yshift=0.65cm]  {(b)} ; %
     \node[const, below=of x, yshift=0.65cm]  {(c)} ; %
    
    % Connect the nodes
    \edge [-] {x} {z} ; %
    \edge [bend left] {xenc} {zenc} ; %
    \edge [bend left] {zdec} {xdec} ; %
    
\end{tikzpicture}
        \phantomcaption
        \label{fig:mim-model}
    \end{subfigure}
    \hfill
    \begin{subfigure}[t]{0.48\textwidth}
        \centering
        \begin{tikzpicture}

    % Define nodes
    
    \node[latent]                                           (zenc) {$\z$};
    \node[obs, below=of zenc, xshift=-1.2cm]                (xenc) {$\x$};
    \node[obs, below=of zenc, xshift=1.2cm]                 (kenc) {$\k$};
    
    \node[latent, right=of zenc, xshift=2.4cm]              (zdec) {$\z$};
    \node[obs, below=of zdec, xshift=-1.2cm]                (xdec) {$\x$};
    \node[obs, below=of zdec, xshift=1.2cm]                 (kdec) {$\k$};
        
     \node[const, below=of xenc, yshift=0.65cm, xshift=1.2cm]  {(d)} ; %
     \node[const, below=of xdec, yshift=0.65cm, xshift=1.2cm]  {(e)} ; %
    
    % Connect the nodes
    \edge  {xenc} {zenc} ; %
    \edge  {zenc} {kenc} ; %
    \edge  {xenc} {kenc} ; %

    \edge  {zdec} {xdec} ; %
    \edge  {zdec}  {kdec} ; %
    \edge  {xdec} {kdec} ; %
    
\end{tikzpicture}
        \phantomcaption
        \label{fig:cmim-model}
    \end{subfigure}

    \caption{(Left) A \MIM model learns two factorizations of a joint distribution:
    (a) encoding; (b) decoding factorizations; and (c) the estimated joint distribution 
    (an undirected graphical model).
    (Right) We extend the \MIM model with an additional binary variable $\k$, and present
    the two factorizations of a joint distribution:
    (c) encoding; (d) decoding factorizations.}
    \label{fig:mim-cmim-overview}
\end{figure}

In this section, we extend the formulation of the Mutual Information Machine (MIM), a probabilistic auto-encoder designed to learn informative and clustered latent codes. The clustering is achieved by minimizing the marginal entropy of the latent distribution over 
$\z$, which results in latent codes that are closely positioned in Euclidean space for similar samples (see example in the work by \cite{reidenbach2023molmim}). In MIM, similarity between samples is defined by the decoding distribution, leading to a local structure around each latent code (\ie, similar samples correspond to nearby latent codes). However, the global distribution of these latent codes, while aligned with a target or learned prior, may not be well-suited for discriminative tasks.
To address this limitation, we propose augmenting the MIM objective with a contrastive objective term, which encourages the latent codes of dissimilar samples to be more distinct from each other. This modification aims to improve the global structure of the latent space, making it more suitable for discriminative downstream tasks.
Throughout the paper, $X$ denotes the observation and $Z$ the latent codes.

%%%%%%%%%%%%%%%%%%%%%%%%%%%%%%%%%%%%%%%%%%%%%%%%%%%%%%%%%%%

\subsection{Contrastive Learning} \label{sec:contrastive-learning}

Contrastive learning is a representation learning technique that maximizes the similarity between positive pairs while minimizing the similarity between negative pairs. The similarity between samples is typically measured using a similarity function, such as cosine similarity $\SIM{\zi}{\zj}=\frac{\zi \cdot \zj}{||\zi||\cdot||\zj||}$. For example, in the InfoNCE loss \cite{oord2018cpc}, the similarity is computed as the dot product of the normalized representations. More formally, we define the similarity of two samples $\xi$ and $\xj$ as $\SIM{\zi}{\zj}$, where the corresponding encoded representations are $\zi = f_\theta(\xi)$, $\zj = f_\theta(\xj)$, and $\EXPSIM{\zi}{\zj} = \exp( \SIM{\zi}{\zj} / \tau )$ to be the exponent of the similarity scaled by scalar temperature $\tau$, and the InfoNCE loss per sample is defined as follows:
\begin{equation}
    \text{InfoNCE}(\xi, \xi^{+}) = - \log \left( \frac{ \EXPSIM{\zi}{\zi^{+}} }{\sum_{j=1}^{B} \EXPSIM{\zi}{\zj} } \right),
    \label{eq:infonce-loss}
\end{equation}
where $\xi^{+}$ is a positive augmentation of the same source sample, and $\xj$ are negative augmentation from different source samples. See \cite{oord2018cpc} for a detailed discussion on the InfoNCE loss.

In practice, the contrastive loss is formulated as a $B$-way classification problem, where the positive pair is distinguished as the first pair, and $B$ represents the batch size. For effective learning, it is crucial that the data augmentation applied to the positive pair is meaningful; however, for certain modalities (\eg, text), designing appropriate augmentations can be challenging. Furthermore, the effectiveness of the loss function is sensitive to the selection of negative examples and the batch size.

%%%%%%%%%%%%%%%%%%%%%%%%%%%%%%%%%%%%%%%%%%%%%%%%%%%%%%%%%%%

\subsection{Contrastive MIM Learning (cMIM)} \label{sec:contrastive-mim}

In this work, we propose augmenting the MIM objective with a contrastive term to introduce global discriminative structure to the locally clustered latent space. Specifically, we hypothesize that encouraging the latent codes of similar samples to be close to each other (\ie, local structure) and the latent codes of dissimilar samples to be distinct (\ie, global structure) are complementary objectives. When comparing MIM with its contrastive extension (cMIM), we expect cMIM to exhibit similar reconstruction fidelity, comparable clustering performance, and improved discriminative capabilities.

Contrastive learning typically relies on generating augmented data for positive pairs, introducing an inductive bias that may not capture all desired invariances within the data. Moreover, devising appropriate augmentations can be challenging for certain data modalities, such as text, as discussed in \cite{lekhac2020contrastive}. Additionally, contrastive learning is sensitive to batch size since it requires a sufficient number of negative examples to improve the quality of learned representations. Although contrastive methods that do not rely on negative examples exist—such as BYOL \cite{grill2020byol}—these methods often introduce additional hyperparameters that can be difficult to tune.

%%%%%%%%%%%%%%%%%%%%%%%%%%%%%%%%%%%%%%%%%%%%%%%%%%%%%%%%%%%

\subsubsection{Contrastive Learning without Data Augmentation} \label{sec:contrastive-learning-without-data-augmentation}

In this work, we propose to extend the MIM framework by introducing a new random variable $\k$ into MIM's graphical model (Fig. \ref{fig:cmim-model}). Formally, the encoding (\ie, $\Menc(\x, \z, \k)$) and decoding (\ie, $\Mdec(\x, \z, \k)$) factorizations of the joint distribution are defined as follows:

\begin{equation}
    \Menc(\x, \z, \k) = \Menc(\k | \x, \z) \; \Menc(\z | \x) \; \Menc(\x)      
\end{equation}

\begin{equation}
    \Mdec(\x, \z, \k) = \Mdec(\k | \x, \z) \; \Mdec(\x | \z) \; \Mdec(\z)      
\end{equation}

Let us first define $\zi$ as a latent code that is sampled from the joint encoding distribution $\Menc(\x=\xi, \z, \k=1)$, where given $\xi$, we sample $\zi \sim \Menc(\z \mid \xi)$. Here we introduced the random variable $\k$, a binary variable representing the relationship between a sample $\xi$ and a latent code $\zj$.
$\k=1$ for the paired sample $(\x,\z)=(\xi,\z_i)$ with $\z_i\!\sim\Menc(\z\mid\xi)$, and $\k=0$ for mismatched pairs $(\xi,\z_j)$, $j\neq i$. As discussed below, introducing $\k$ allows the model to learn a contrastive objective without relying on data augmentation, which is typically required in contrastive learning. Choosing directional similarity (cosine) as the similarity function further encourages dissimilar samples to diverge in angle, while MIM's original loss encourages similar samples to cluster in Euclidean space.

To elaborate, we define the discriminator distributions for the encoding and decoding factorizations over $\k$ as:

\begin{equation}
    \Menc(\k \mid \z = \zi, \x) = \Mdec(\k \mid \z = \zi, \x) = \text{Bernoulli}(\k; p_{k=1}),
\end{equation}

where

\begin{equation}
    \begin{aligned}
        p_{k=1}(\xi, \zi) &= \frac{\EXPSIM{\zi}{\zi}}{\EXPSIM{\zi}{\zi} + \E{\x' \sim \pjoint(\x),\z' \sim \Menc(\z | \x')}{ \EXPSIM{\zi}{\z'}} } \\
        &\approx \frac{\EXPSIM{\zi}{\zi}}{\EXPSIM{\zi}{\zi} + \frac{1}{B-1} \sum_{\substack{j=1 \\ j \neq i}}^B \EXPSIM{\zi}{\zj} } \label{eq:pk1}
    \end{aligned}
\end{equation}
where $\EXPSIM{\cdot}{\cdot}$ is the exponentiated logits defined in Section \ref{sec:contrastive-learning}, and $\SIM{\cdot}{\cdot}$ is cosine similarity, and with $\x' \neq \xi$ when using in-batch approximation.
We emphasize that both encoding and decoding distributions are identical, to further encourage consistency between the encoding and decoding distributions, in addition to the consistency promoted by MIM's loss.

During training we always have $\k = 1$ since $\zi$ is sampled from the encoding distribution given $\xi$. The expectation over other samples is approximated using the current batch of size $B$, excluding the current sample $\xi$. This approach leads to simpler training procedure when compared to InfoNCE, and allows us to avoid the need for positive data augmentation, as we can directly compare the latent code $\zi$ with latent codes from other samples in the batch. See Algo. \ref{algo:cmim} for the training procedure of cMIM.

A key advantage of cMIM is reduced sensitivity to batch size. Moreover, since the sampling process guarantees $\k=1$, the model is never trained with $\k=0$ samples, leading to a simplified empirical loss. By formulating the objective in terms of expectations—rather than a $B$-way classification as in standard contrastive loss—the expected similarity to other samples can be approximated efficiently via Monte Carlo sampling. This, in principle, decouples similarity estimation from batch size, making the method less sensitive to batch size as it grows, consistent with bounds from Hoeffding’s inequality \cite{Hoeffding:1963}.

Because our similarity $\SIM{\cdot}{\cdot}$ is bounded (cosine similarity in $[-1,1]$ implies
$\EXPSIM{\cdot}{\cdot}\in[\mathrm e^{-1/\tau},\mathrm e^{1/\tau}]$), the in-batch Monte Carlo
estimate concentrates around its expectation. In particular, Hoeffding’s inequality yields
\begin{equation}
\Pr\!\left(\left|\tfrac{1}{B-1}\!\sum_{j\ne i}\EXPSIM{\z_i}{\z_j}-\mu\right|\ge \epsilon\right)
\le 2\exp\!\left(-\frac{2(B-1)\epsilon^2}{(\mathrm e^{1/\tau}-\mathrm e^{-1/\tau})^2}\right).
\end{equation}
Thus the estimator becomes more stable as $B$ increases (variance $\mathcal{O}(1/(B{-}1))$) \cite{Hoeffding:1963}.

Another simplification is that cMIM requires no data augmentation, unlike conventional contrastive learning where augmentations are essential for defining positive pairs. In cMIM, clustering is already promoted by the MIM objective, removing this requirement. This avoids the often challenging design and tuning of augmentations, further reducing training complexity. The combination of clustered latent codes with discriminative angular structure yields a latent space with stronger discriminative structure.

%%%%%%%%%%%%%%%%%%%%%%%%%%%%%%%%%%%%%%%%%%%%%%%%%%%%%%%%%%%

\subsubsection{Contrastive MIM and InfoNCE} \label{sec:cmim-infonce}

To better understand the proposed contrastive loss, we can relate it to the InfoNCE loss. In InfoNCE, the positive pair is defined by data augmentation, encourages the model to maximize the similarity between augmented samples and minimize the similarity for other samples. In contrast, MIM already clusters similar samples in latent space, and the contrastive term is used to encourage dissimilar samples to be more distinct from each other. We can rewrite Eq. \eqref{eq:pk1} as follows:

\paragraph{Relation to InfoNCE}

Let $s_{ij} \!\triangleq\! \SIM{\zi}{\zj}/\tau$ so that
$\EXPSIM{\z_i}{\z_j}=\exp(s_{ij})$.
From Eq.~\eqref{eq:pk1} we have
\begin{equation}
    \begin{aligned}
p_{k=1} &= \frac{\exp(s_{ii})}{\exp(s_{ii})+\tfrac{1}{B-1}\sum_{j\neq i}\exp(s_{ij})} \\
        &= \frac{(B-1)\exp(s_{ii})}{(B-1)\exp(s_{ii})+\sum_{j\neq i}\exp(s_{ij})} \\
        &= \frac{\exp\!\big(s_{ii}+\log(B{-}1)\big)}
       {\exp\!\big(s_{ii}+\log(B{-}1)\big)+\sum_{j\neq i}\exp(s_{ij})}.
    \end{aligned}
\end{equation}

\noindent\textbf{Proposition.}
The probability $p_{k=1}$ above equals the softmax over the $B$ candidates
$\{\z_i,\{\z_{j\neq i}\}\}$ where the \emph{positive} logit is shifted by a constant
$\log(B{-}1)$:
\[
p_{k=1} \;=\; \operatorname{softmax}\!\Big([\,s_{ii}{+}\log(B{-}1),~\{s_{ij}\}_{j\neq i}\,]\Big)_{\text{pos}}.
\]
Consequently, minimizing $-\log p_{k=1}$ is identical to an InfoNCE cross‑entropy
computed on logits $\{s_{ii}{+}\log(B{-}1),\,s_{ij}\ (j\neq i)\}$, \ie, \emph{InfoNCE with a
fixed positive‑logit offset}.

\noindent\textbf{Remarks.}
(i) \emph{Calibration.} If all logits are equal ($s_{ij}\!=\!s_{ii}$), then
$p_{k=1}=1/2$ (independent of $B$), whereas standard InfoNCE (unshifted softmax with
$\sum_{j=1}^{B}\exp(s_{ij})$ in the denominator) yields $p_{k=1}=1/B$.
(ii) \emph{Gradient shape.} With cosine similarity, $s_{ii}=1/\tau$ is constant, so the
positive contributes no gradient and the gradients originate solely from the negative mean;
local attraction (\ie, clustering) is provided by the MIM term.
(iii) \emph{Exact InfoNCE and MI bound.} If the mean over negatives in
Eq.~\eqref{eq:pk1} is replaced by the \emph{sum} (\ie, drop the factor $1/(B{-}1)$), the
offset disappears and one recovers standard InfoNCE
$\frac{\exp(s_{ii})}{\sum_{j=1}^{B}\exp(s_{ij})}$ together with the classical
$\,I(X;Z)\!\ge\!\log B-\mathbb{E}[\mathcal{L}_\text{InfoNCE}]$ bound.
We point the reader to \cite{oord2018cpc} for a detailed discussion on the optimum of the InfoNCE loss.

\paragraph{Differences from InfoNCE}

Since cMIM does not rely on data augmentation, it does not require the positive pair to be defined by augmentations. Instead, the positive pair is defined by the latent code $\zi$ sampled from the encoding distribution given $\xi$. This allows cMIM to avoid the challenges associated with designing appropriate augmentations for different modalities, such as text, where augmentations can be difficult to define.

In addition, our mean-denominator form calibrates the positive probability at $p_{k=1}=1/2$
when logits are equal (independent of $B$), and the rapid convergence of the expectation estimator 
makes cMIM more robust to variations in batch size, as it does not strongly rely on the number of negative samples.
We support this claim with empirical results in Section \ref{sec:experiments}. 
We also note that the different calibration prevents cMIM from enjoying the classical InfoNCE mutual information bound.
However, cMIM does inherit MIM's mutual information bound (see \cite{livne2019mim}), which is sufficient for learning informative latent codes.

That said, the quality of the Monte Carlo estimator
and the diversity of negatives still improve with larger batches (or a memory queue).
Empirically, we observe reduced calibration sensitivity but still a monotonic gain
from increasing the effective number of negatives.

%%%%%%%%%%%%%%%%%%%%%%%%%%%%%%%%%%%%%%%%%%%%%%%%%%%%%%%%%%%

\subsubsection{cMIM Training Procedure} \label{sec:cmim-training-algorithm}

\begin{figure}[t]
    \centering
    \begin{minipage}[t]{0.95\columnwidth}
    \begin{algorithm}[H]
        \small
        \caption{Learning parameters $\params$ of cMIM}
        \label{algo:cmim}
        \begin{algorithmic}[1]
            \REQUIRE Samples from dataset $\pjoint(\x)$
            \WHILE{not converged}
            \STATE $\mathcal{D} \gets \{ \x_j, \z_j \sim \Menc(\z|\x)\pjoint(\x) \}_{j=1}^{B}$ \COMMENT{\textcolor{gray}{\textit{Sample a batch of size $B$}}}
            \STATE $\EAMIMloss \left( \params ; \mathcal{D} \right) = -\frac{1}{B}\! \sum_{i=1}^{B}\! \big( ~ \log \Mdec(\x_i | \z_i) + \log p_{k=1}(\x_i, \z_i)$
            \STATE \quad\quad\quad\quad\quad\quad\quad\quad\quad\quad\quad\quad$+ \frac{1}{2} \left( \log \Menc(\z_i | \x_i) + \log \pjoint(\z_i) \right) ~ \big)$
            \STATE $\Delta \params \propto -\nabla_{\params}  \EAMIMloss \left( \params ; \mathcal{D} \right)$
            \COMMENT{\textcolor{gray}{\textit{Gradient computed through sampling using reparameterization}}}
            \ENDWHILE
        \end{algorithmic}
    \end{algorithm}
    \end{minipage}
    \caption{Training algorithm for cMIM.}
    \end{figure}

The training of cMIM is conducted using the MIM objective applied to the extended graphical model, as discussed in \cite{livne2019mim}. Specifically, MIM is defined over a mixture model as follows:
\begin{equation}
    \Mmodel (\x, \z, \k) = \frac{1}{2} \left( \Mdec(\k \mid \z, \x)\, \Mdec(\x \mid \z)\, \Mdec(\z) + \Menc(\k \mid \z, \x)\, \Menc(\z \mid \x)\, \Menc(\x) \right),
\end{equation}
with a sampling distribution $\Msamp(\x, \z, \k)$, given by
\begin{equation}
    \Msamp(\x, \z, \k) = \Pmix,
\end{equation}
as discussed in \cite{livne2019mim}, where we introduce here the discriminator distributions over $\k$.

The learning process for MIM involves minimizing an upper bound, defined as follows:
\begin{equation}
    \begin{aligned}
        \MIMloss(\params) = &
        \frac{1}{2} \Big(\, \CE{\Msamp(\x, \z, \k)}{\Menc \left(\x, \z, \k \right)} \\
        &\quad + ~ \CE{\Msamp(\x, \z, \k)}{\Mdec \left(\x, \z, \k \right)} \, \Big) \\
        &\ge H_{\Msamp} (\x, \k) + H_{\Msamp} (\z)  - I_{\Msamp} (\x, \k;\z)
    \end{aligned}
\end{equation}
where we treat $\k, \x$ as a single observed variable during training, and remind the reader that $\k = 1$ for all samples $\x$ \cite{livne2019mim}.

We also provide an explicit example of the loss for A-MIM, an asymmetric version of MIM, as discussed in \cite{livne2019mim}.
In A-MIM we sample $(\x,\z,\k)$ from the encoding path but evaluate cross-entropies under both factorizations:
\begin{equation}
    \AMIMloss (\params) = -\frac{1}{2} \E{\x \sim \pjoint(\x),\z \sim \Menc(\z|\x), \k = 1}{ 
        \begin{aligned}
            \log \Mdec(\k | \z, \x) + \log & \Mdec(\x | \z) + \log \Mdec(\z) \\
            &+ \\
            \log \Menc(\k | \z, \x) + \log & \Menc(\z | \x) + \log \Menc(\x)
        \end{aligned}
    } \label{eq:mim-loss}
\end{equation}
where we extend the loss from \cite{livne2019mim} by incorporating $\k$ into the joint distribution.

The training algorithm for cMIM is presented in Algorithm \ref{algo:cmim}, where we extend the learning procedure from \cite{livne2020sentencemim}. The final empirical loss is defined below:
\begin{equation}
\begin{aligned}
\EAMIMloss \left( \params ; \mathcal{D} \right) &= -\frac{1}{N}\! \sum_{i=1}^{N}\! \big( ~ \log \Mdec(\x_i | \z_i) + \log p_{k=1}(\x_i, \z_i) \\
&\quad + \frac{1}{2} \left( \log \Menc(\z_i | \x_i) + \log \pjoint(\z_i) \right) ~ \big)
\end{aligned}
\end{equation}
where $\mathcal{D} = \{\xi,\zi\ \sim \Menc(\z|\x) \pjoint(\x) \}_{i=1}^{N}$ is the sample set, $p_{k=1}$ is defined in Eq. \eqref{eq:pk1}, $p_{k=1}$ is used for both the encoding and decoding distributions, and the anchor $\pjoint(\z_i)=\mathcal{N}(\z=\zi|\mu=0, \sigma=1)$ is a diagonal Normal distribution.

%%%%%%%%%%%%%%%%%%%%%%%%%%%%%%%%%%%%%%%%%%%%%%%%%%%%%%%%%%%

\subsection{Informative Embeddings} \label{sec:informative-embeddings}

\begin{figure}[t]
    \centering
    \setlength{\tabcolsep}{0pt}
    \includegraphics[width=1.0\columnwidth]{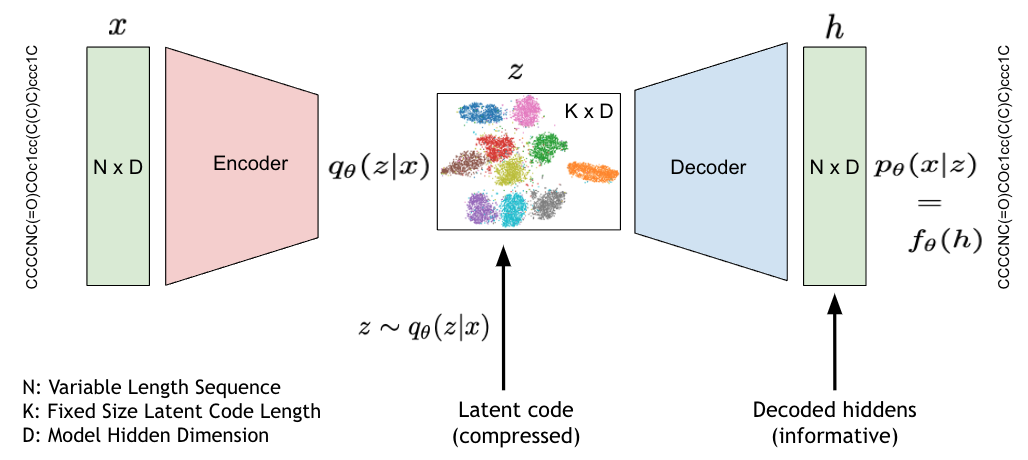}
    \caption{Informative embeddings $\h$ are extracted from an input observation $\x$ by taking the output from the decoder's hidden states, prior to their mapping to the parameters of the decoding distribution $\Mdec$. Teacher forcing should be applied when necessary, particularly in the case of auto-regressive distributions.
    }
    \label{fig:informative-embeddings}
\end{figure}

Inspired by the success of leveraging output hidden states from various layers (\eg, \cite{zeng2024similardatapointsidentification}) of GPT-based large language models (LLMs, see \cite{brown2020gpt3}) for downstream tasks,
and recognizing the limitations of using such hidden states without further fine-tuning (\eg, NV-Embed by \cite{lee2024nv}),
we propose a generic method to enhance the quality of embeddings extracted from the MIM model, as depicted in Fig. \ref{fig:informative-embeddings}.

Specifically, the embeddings $\h$ are extracted from the hidden states of the decoder, just before they are mapped to the parameters of the decoding distribution $\Mdec(\x \mid \z) = f_{\params}(\h)$. These embeddings are then utilized for various downstream tasks, including classification, clustering, and generation. 

For auto-regressive distributions, such as text, teacher forcing is employed, where the input $\x$ is fed into both the encoder and decoder, eliminating the need to generate the next token. For non-auto-regressive distributions, such as images, there is no need for teacher forcing, and the embeddings $\h$ are directly used for tasks like classification or clustering. More formally, the embeddings $\hi$ for sample $\xi$ are defined as:
\begin{equation}
    \hi = \text{Decoder}(\xi | \zi \sim \Menc(\z | \xi)) = \text{Decoder}(\xi, \text{Encoder}(\xi)),
\end{equation}
where the above equation is employing teacher forcing for auto-regressive distributions.

We propose that instead of directly using the latent codes from MIM for downstream tasks, the focus should be on the embeddings $\h$ derived from the decoder's hidden states. As we demonstrate, these embeddings are crucial for downstream tasks and are adapted according to the specific characteristics of the data, whether auto-regressive or non-auto-regressive. We note that informative embeddings can be extracted from any encoder-decoder model, not just MIM.

To illustrate the intuition behind using informative embeddings, consider the case of text generation. The embeddings $\h$, extracted from the decoder's hidden states by averaging over the sequence dimension, encapsulate information from the latent code that has been transformed to represent the probability distribution of the next token. This can be viewed as an "enriched" representation of the latent code, augmented with additional contextual information from the decoder, resulting in a more comprehensive representation of the observation (\ie, capturing the entire distribution rather than just the actual observation). By leveraging these enriched embeddings for downstream tasks, we can exploit the additional information contained in the decoder's hidden states, potentially enhancing performance in discriminative tasks such as classification or regression.
Unless stated otherwise, for sequences we mean-pool decoder hidden states to obtain $\h_i\in\mathbb{R}^d$; gradients flow through both encoder and decoder during cMIM training.

\section{Experiments} \label{sec:experiments}

To evaluate the proposed cMIM model, we conduct experiments on a 2D toy example, MNIST-like images, and on molecular property prediction tasks (MolMIM by \cite{reidenbach2023molmim}).
The 2D toy example illustrates the impact of the proposed contrastive MIM loss.
We then explore the qualitative nature of cMIM by running rigorous experiments on MNIST-like datasets.
Finally, we compare the performance of cMIM with MIM, VAE, and AutoEncoder models trained on molecular data, assessing their reconstruction and effectiveness in downstream tasks. 

%%%%%%%%%%%%%%%%%%%%%%%%%%%%%%%%%%%%%%%%%%%%%%%%%%%%%%%%%%%
\subsection{Experiment Details and Datasets}

\begin{table}[ht]
\centering
\begin{tabular}{clcccl}
\hline
\textbf{\#} & \textbf{Dataset} & \textbf{Train Samples} & \textbf{Test Samples} & \textbf{Categories} & \textbf{Description} \\
\hline
1  & MNIST          & 60,000   & 10,000    & 10  & Handwritten digits \\
2  & Fashion MNIST  & 60,000   & 10,000    & 10  & Clothing images \\
3  & EMNIST Letters & 88,800   & 14,800    & 27  & Handwritten letters \\
4  & EMNIST Digits  & 240,000  & 40,000    & 10  & Handwritten digits \\
5  & PathMNIST      & 89,996   & 7,180     & 9   & Colon tissue histology \\
6  & DermaMNIST     & 7,007    & 2,003     & 7   & Skin lesion images \\
7  & OCTMNIST       & 97,477   & 8,646     & 4   & Retinal OCT images \\
8  & PneumoniaMNIST & 9,728    & 2,433     & 2   & Pneumonia chest X-rays \\
9  & RetinaMNIST    & 1,600    & 400       & 5   & Retinal fundus images \\
10 & BreastMNIST    & 7,000    & 2,000     & 2   & Breast tumor ultrasound \\
11 & BloodMNIST     & 11,959   & 3,432     & 8   & Blood cell microscopy \\
12 & TissueMNIST    & 165,466  & 47,711    & 8   & Kidney tissue cells \\
13 & OrganAMNIST    & 34,581   & 8,336     & 11  & Abdominal organ CT scans \\
14 & OrganCMNIST    & 13,000   & 3,239     & 11  & Organ CT, central slices \\
15 & OrganSMNIST    & 23,000   & 5,749     & 11  & Organ CT, sagittal slices \\
\hline
\end{tabular}
\vspace{1.0em}
\caption{\textbf{Image Classification:} Summary of train/test samples, categories, and descriptions for MNIST, EMNIST, and MedMNIST datasets.}
\label{tab:mnist-datasets}
\end{table}

All models are trained in an unsupervised manner.
The checkpoint with the lowest validation loss is selected for evaluation.
We avoid selecting any intermediate checkpoints, a common heuristic which does not scale well with complexity and model size. 
For downstream tasks, we freeze the encoder--decoder and train lightweight classifiers on the learned representations, evaluating them on the held-out test set.
Importantly, classification accuracy is not monitored during training.
This ensures that comparisons reflect the quality of unsupervised representations rather than reliance on checkpoint selection heuristics.
Training continues until convergence, making comparisons fair across models.

\paragraph{2D Toy Example.}
A synthetic dataset of 1000 points in 2D space is initialized in the first quadrant.
The task is to check the effect of the proposed contrastive MIM loss on their position, highlighting the effect of Eq. \eqref{eq:pk1}.

\paragraph{Image Classification on MNIST-like Datasets.}
We train MIM, cMIM, VAE, cVAE (VAE with the contrastive term), and InfoNCE to convergence on MNIST-like datasets.
Comparisons include classification accuracy, batch size sensitivity, and reconstruction (except InfoNCE).
Datasets include MNIST \cite{6296535}, EMNIST (letters, digits) \cite{DBLP:journals/corr/CohenATS17}, and MedMNIST \cite{DBLP:journals/corr/abs-2110-14795}.
All images are resized to $28 \times 28$ pixels and converted to black and white when needed.
We used $\tau = 0.1$ (as in InfoNCE) following a small hyper-parameter search with $\tau \in \{0.1,1\}$.
The encoder is a Perceiver \cite{jaegle2021perceiver} with 1 cross-attention layer, 4 self-attention layers, hidden size 16, projecting 784 pixels to 400 steps, followed by a projection to 64-dimensional latent codes.
The decoder mirrors this design.
Models are trained for 500k steps with batch sizes 2, 5, 10, 100, 200, using Adam with learning rate $10^{-3}$, and WSD scheduler \cite{hu2024minicpm}.
Classifiers include KNN ($k=5$; cosine and Euclidean metrics) and a one-hidden-layer MLP (size 400, Adam, $10^{-3}$, 1000 steps).

\paragraph{Molecular Property Prediction.}
We use ZINC-15 \cite{Sterling2015zinc15} with SMILES \cite{Weininger1988smiles} sequences, following \cite{reidenbach2023molmim}.
Properties include ESOL, FreeSolv, and Lipophilicity.
Here $\tau = 1$.
Both MIM and cMIM are trained for 250k steps.
Embeddings are evaluated using SVM and MLP regressors, both with and without informative embeddings, and compared against CDDD \cite{C8SC04175J}.
Architectural details appear in Appendix \ref{sec:appendix-model-arch}.

%%%%%%%%%%%%%%%%%%%%%%%%%%%%%%%%%%%%%%%%%%%%%%%%%%%%%%%%%%%

\subsection{Effects of cMIM Loss on 2D Toy Example}

\begin{figure}[t]
    \centering
    \begin{tabular}{@{}c@{}c@{}c@{}c@{}}
        \begin{tabular}{c}
            \includegraphics[width=0.22\textwidth,trim=70pt 0 70pt 0,clip]{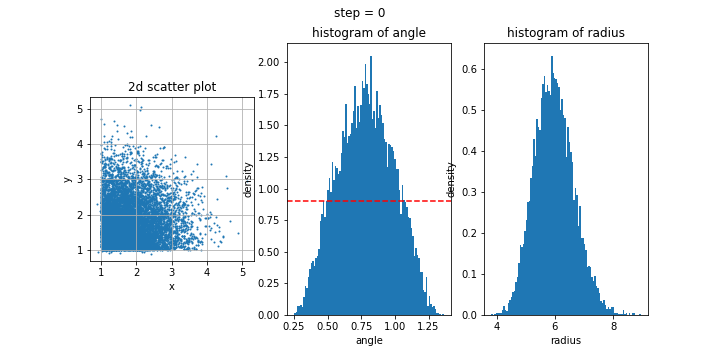} \\
            \textbf{(a)} Step 0
        \end{tabular} &
        \begin{tabular}{c}
            \includegraphics[width=0.22\textwidth,trim=70pt 0 70pt 0,clip]{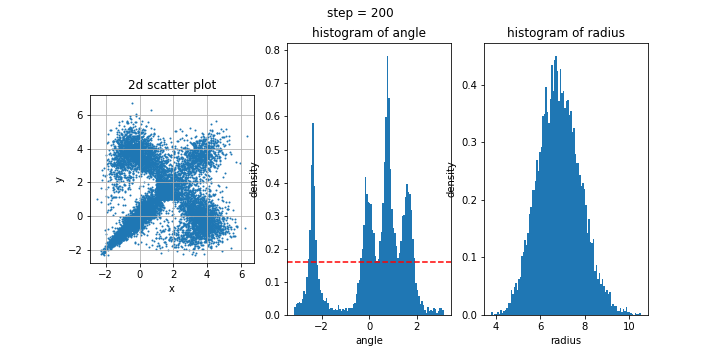} \\
            \textbf{(b)} Step 200
        \end{tabular} &
        \begin{tabular}{c}
            \includegraphics[width=0.22\textwidth,trim=70pt 0 70pt 0,clip]{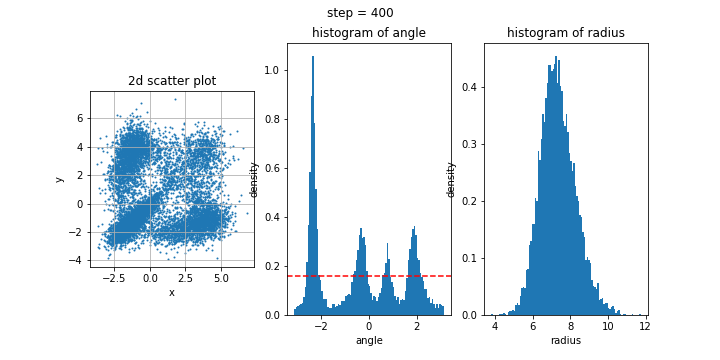} \\
            \textbf{(c)} Step 400
        \end{tabular} &
        \begin{tabular}{c}
            \includegraphics[width=0.22\textwidth,trim=70pt 0 70pt 0,clip]{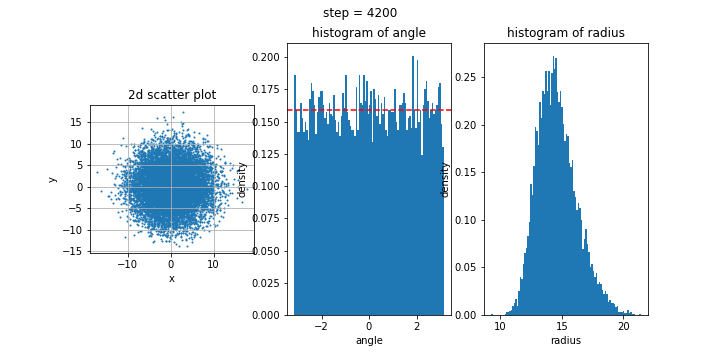} \\
            \textbf{(d)} Step 4200
        \end{tabular} 
    \end{tabular}
    \caption{Effect of the contrastive MIM loss term in Eq. \eqref{eq:pk1} on the 2D toy example. Each plot shows: latent space (left), histogram of latent code angles (middle), and histogram of latent radii (right). From \textbf{(a)} initialization in the first quadrant to \textbf{(d)} after 4200 training steps, the loss distributes points uniformly in angle while allowing radii to vary. This complements MIM’s clustering by encouraging angular uniformity, which enhances separability and thus improves downstream discriminative performance.}
    \label{fig:cMIM-to-2d}
\end{figure}

We study the effect of the proposed contrastive MIM loss term (Eq. \eqref{eq:pk1}) on a 2D toy example.
Training minimizes the negative log-likelihood associated with Eq. \eqref{eq:pk1}, learning optimal latent codes in two dimensions.
Here we use $\tau = 1$.

We expect the latent codes to distribute uniformly across all angles while maintaining variability in radii, as suggested by \cite{wang2020understanding}.
Fig. \ref{fig:cMIM-to-2d} shows the progression of the latent space during training.
The results confirm that the contrastive term integrates smoothly with the MIM objective, preserving radial clustering while enforcing uniform angular distribution.

%%%%%%%%%%%%%%%%%%%%%%%%%%%%%%%%%%%%%%%%%%%%%%%%%%%%%%%%%%%

\subsection{Classification Accuracy}

\begin{figure}[t]
    \centering
    \begin{tabular}{cc}
        \includegraphics[width=0.48\textwidth]{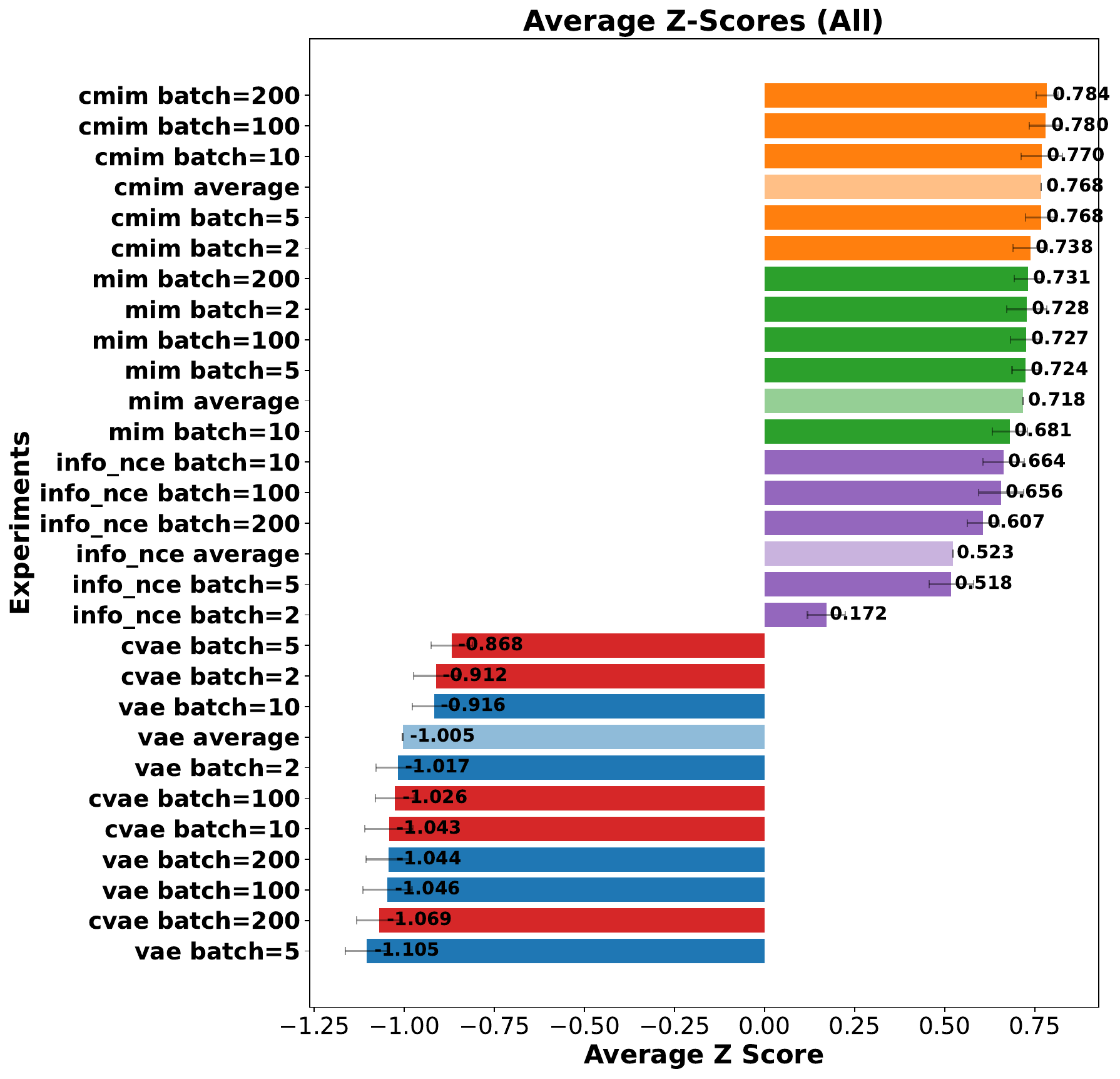} &
        \includegraphics[width=0.48\textwidth]{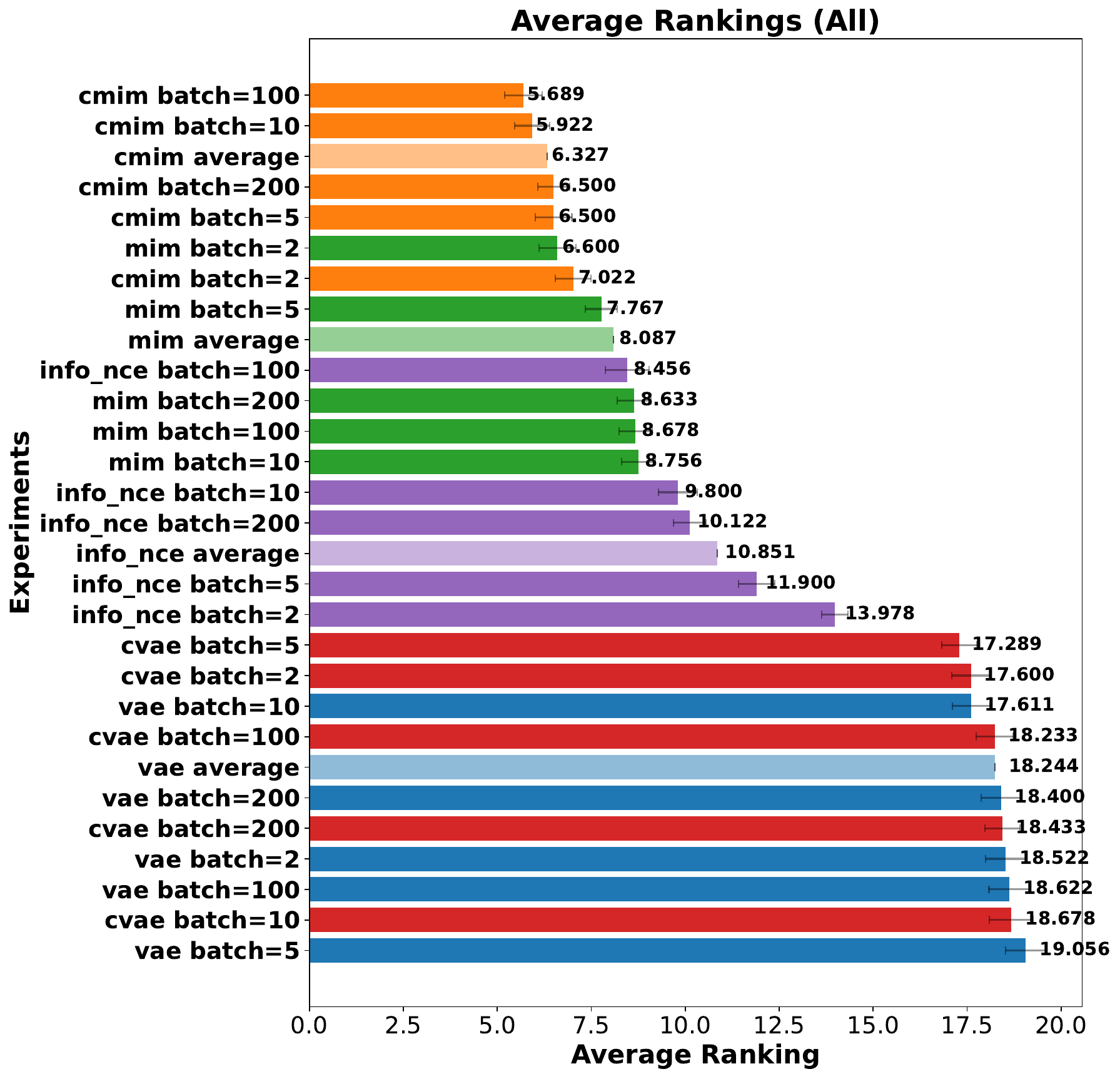} \\
        \textbf{(a)} Z-scores with error bars & \textbf{(b)} Rankings with error bars
    \end{tabular}
    \caption{Classification accuracy across datasets and classifiers. Colors indicate model families: cMIM (orange), MIM (green), InfoNCE (purple), VAE (blue), cVAE (red). Light shades denote model averages. cMIM consistently outperforms all baselines across batch sizes and metrics.}
    \label{fig:mnist-classification-accuracy}
\end{figure}

\begin{table}[t]
    \setlength{\tabcolsep}{5pt}
    \renewcommand{\arraystretch}{1.5}
    \centering
    \begin{tabular}{l||c|c||c|c||c|c||c}
        \hline
        \multirow{2}{*}{\textbf{Model (Latent K $\times$ D)}} & \multicolumn{2}{c||}{\textbf{ESOL}} & \multicolumn{2}{c||}{\textbf{FreeSolv}} & \multicolumn{2}{c||}{\textbf{Lipophilicity}} & \textbf{Recon.} \\
        \cline{2-7} 
        & \textbf{SVM} & \textbf{MLP} & \textbf{SVM} & \textbf{MLP} & \textbf{SVM} & \textbf{MLP} & \\
        \hline\hline
        MIM (1 $\times$ 512) & 0.65 & 0.34 & 2.23 & 1.82 & 0.663 & 0.61 & 100\% \\
        cMIM (1 $\times$ 512) & 0.47 & \cellcolor{yellow!25}0.19 & 2.32 & 1.67 & 0.546 & 0.38 & 100\% \\
        MIM (1 $\times$ 512) info emb & 0.21 & 0.29 & 1.55 & 1.4 & 0.234 & 0.28 & 100\% \\
        cMIM (1 $\times$ 512) info emb & \cellcolor{yellow!25}0.21 & 0.24 & 1.74 & \cellcolor{yellow!25}1.35 & 0.24 & \cellcolor{yellow!25}0.23 & 100\% \\
        \hline\hline
        CDDD (512) & \textbf{0.33} &  & \textbf{0.94} &  & \textbf{0.4} &  &  \\
        \textdagger Seq2seq (N $\times$ 512) & 0.37 & 0.43 & 1.24 & 1.4 & 0.46 & 0.61 & 100\% \\
        \textdagger Perceiver (4 $\times$ 512) & 0.4 & 0.36 & 1.22 & 1.05 & 0.48 & 0.47 & 100\% \\
        \textdagger VAE (4 $\times$ 512) & 0.55 & 0.49 & 1.65 & 3.3 & 0.63 & 0.55 & 46\% \\
        MIM (1 $\times$ 512) & 0.58 & 0.54 & 1.95 & 1.9 & 0.66 & 0.62 & 100\% \\
        \hline\hline
        Morgan fingerprints (512) & 1.52 & 1.26 & 5.09 & 3.94 & 0.63 & 0.61 &  \\
    \end{tabular}
    \vspace{1.0em}
    \caption{Comparison of models on ESOL, FreeSolv, and Lipophilicity using SVM and MLP regressors, with reconstruction accuracy. Top: our results. Bottom: results from \cite{reidenbach2023molmim}. For \textdagger models, sequence representations were averaged to 512 dimensions. Bold: best non-MIM results. Highlighted: best among MIM-based models. Note that CDDD training included these classification tasks.}
    \label{tab:cmim-comparison-molmim}
\end{table}

We now analyze classification accuracy, as a proxy for the quality of the learned embeddings, and which was never used as a training signal.

\paragraph{Image Classification.}
We trained MIM, cMIM, VAE, cVAE (VAE + cMIM contrastive loss term), and InfoNCE across batch sizes $\{2,5,10,100,200\}$.
All models share the same architecture, with InfoNCE consisting only of the encoder.
Checkpoints with the lowest validation loss were evaluated on test sets.
This design controls for architecture, optimizer, training steps, and dataset usage, isolating the effect of the objective.

We report results using KNN (cosine and Euclidean) and a one-hidden-layer MLP with 400 dimensions. We use Scikit-learn \cite{JMLR:v12:pedregosa11a} with default values.
Inputs to classifiers are either the mean encoding (standard embedding) or informative embeddings (Section \ref{sec:informative-embeddings}). Together we evaluate 6 classification tasks per model and batch size.
Performance is summarized with (1) average normalized accuracy (z-score across datasets and tasks) and (2) average ranking per dataset and task (Fig. \ref{fig:mnist-classification-accuracy}).

cMIM consistently outperformed all baselines across batch sizes and metrics, dominating the ranking plots.
Both MIM and cMIM surpassed VAE, cVAE, and most InfoNCE runs in z-score.
The only competitive non-MIM model was InfoNCE with batch size 100.
Adding a contrastive term to VAE (cVAE) had little impact, indicating that cMIM’s advantage arises from combining MIM’s clustering with angular separation.

\paragraph{Molecular Property Prediction and Informative Embeddings.}
Table \ref{tab:cmim-comparison-molmim} compares MIM and cMIM on ESOL, FreeSolv, and Lipophilicity tasks.
We evaluate SVM and MLP regressors trained on embeddings and informative embeddings.
Baselines include CDDD \cite{C8SC04175J}, Seq2seq, Perceiver, VAE, and Morgan fingerprints. We note that CDDD was trained with the classification tasks during training.

cMIM with informative embeddings outperformed vanilla MIM and was competitive with or superior to these baselines.
This highlights the value of informative embeddings and the discriminative structure encouraged by cMIM.

%%%%%%%%%%%%%%%%%%%%%%%%%%%%%%%%%%%%%%%%%%%%%%%%%%%%%%%%%%%

\subsection{Batch Size Sensitivity}

\begin{wrapfigure}{r}{0.5\textwidth}
    \centering
    \includegraphics[width=0.48\textwidth]{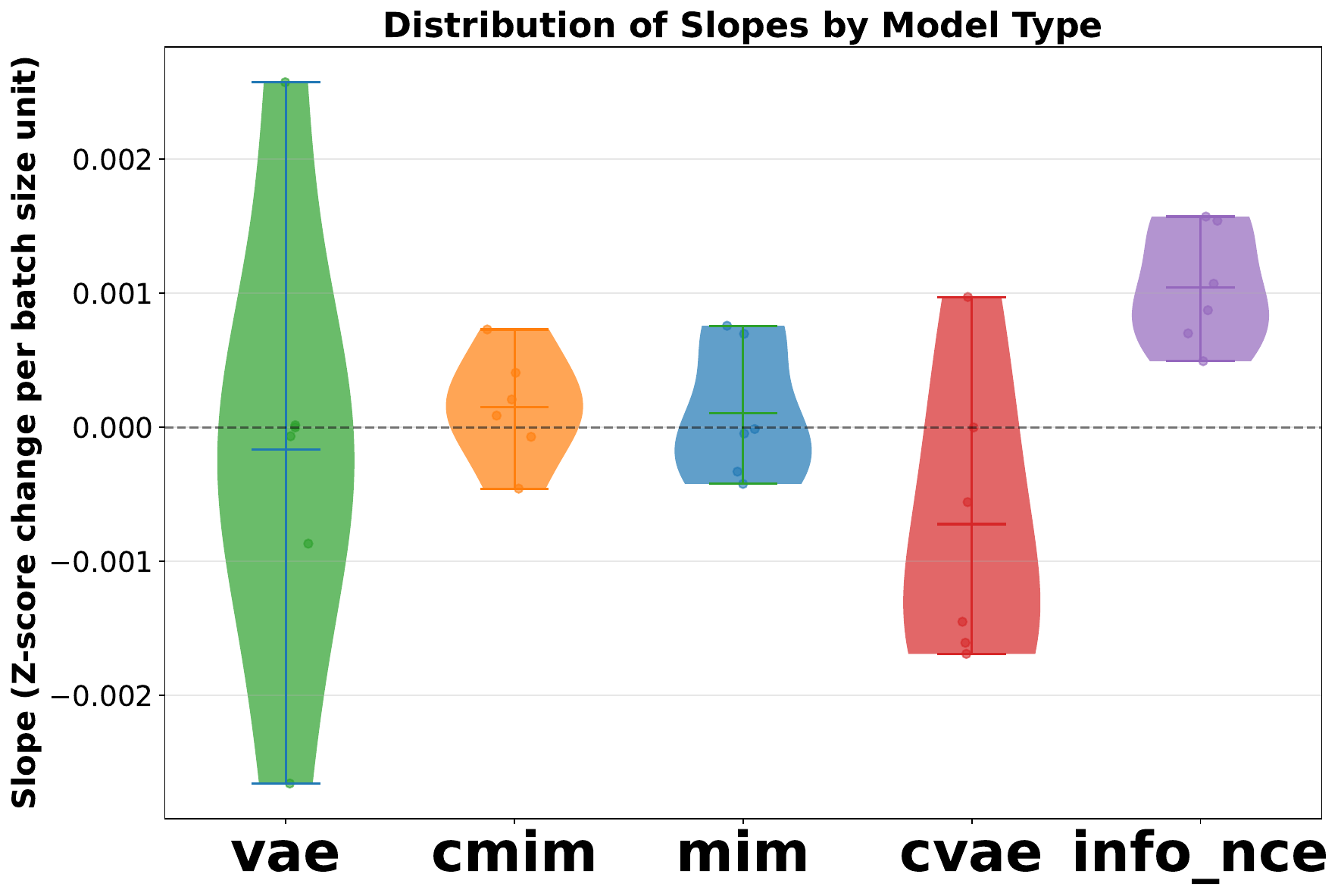}
    \caption{Distribution of slopes from linear fits of accuracy vs. batch size for different models. Each point corresponds to average z-score of a model trained on MNIST-like datasets.}
    \label{fig:batch-size-sensitivity}
\end{wrapfigure}

Fig. \ref{fig:batch-size-sensitivity} summarizes batch size sensitivity for MIM, cMIM, VAE, cVAE, and InfoNCE.
For each model, we performed a linear regression of average z-score on batch size, across six evaluation settings (three classifiers $\times$ two embedding types).
The slope of this fit serves as a measure of sensitivity: positive slope means higher accuracy with larger batches, while near-zero slope indicates robustness.

InfoNCE shows clear dependence on batch size, with positive slopes and low variance.
MIM and cMIM both yield slopes near zero with small variance, confirming their robustness to batch size.
By contrast, VAE and cVAE exhibit high variance in slopes, reflecting unstable performance and strong sensitivity to batch size changes.

%%%%%%%%%%%%%%%%%%%%%%%%%%%%%%%%%%%%%%%%%%%%%%%%%%%%%%%%%%%

\subsection{Reconstruction}

\begin{figure}[t]
    \centering
    \begin{tabular}{cc}
        \includegraphics[width=0.48\textwidth]{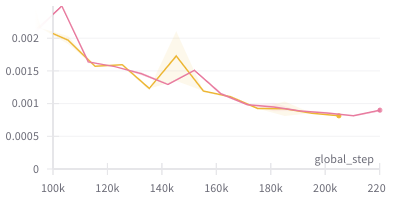} &
        \includegraphics[width=0.48\textwidth]{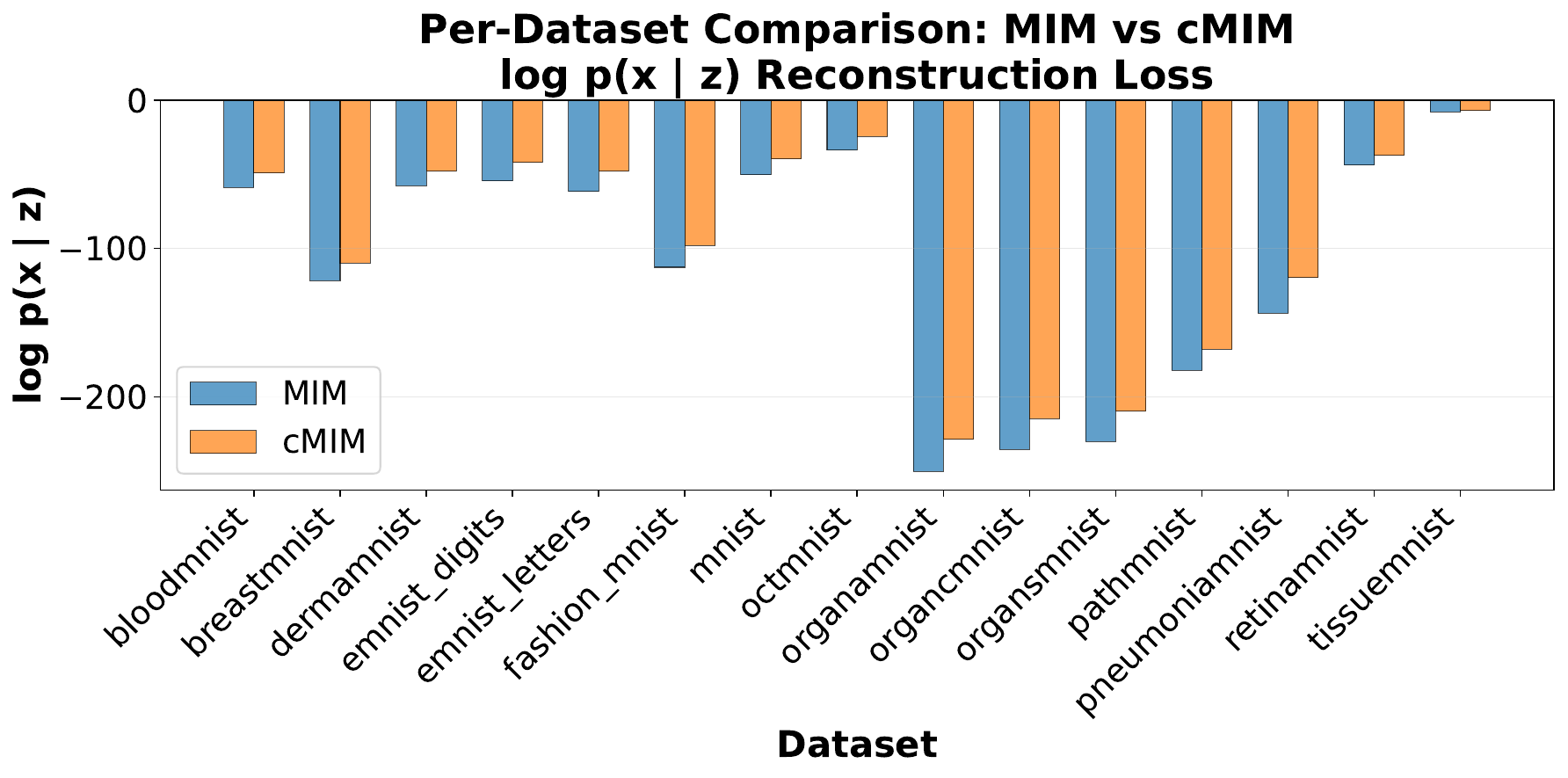} \\
        \textbf{(a)} ZINC15 Training Validation Loss & \textbf{(b)} MNIST Dataset Comparison
    \end{tabular}
    \caption{Reconstruction performance of MIM and cMIM. \textbf{(a)} Validation reconstruction loss during training on molecular data (cMIM in yellow, MIM in pink) shows comparable behavior. \textbf{(b)} Per-dataset test reconstruction log-likelihood on MNIST-like datasets. Surprisingly, cMIM achieves better average reconstruction (-96.25) compared to MIM (-109.64), a +12.2\% relative improvement, suggesting a beneficial regularization effect.}
    \label{fig:cMIM-vs-MIM-WB}
\end{figure}

Fig. \ref{fig:cMIM-vs-MIM-WB} compares reconstruction quality of MIM and cMIM.
On molecular data (panel \textbf{a}), both models exhibit nearly identical reconstruction loss trajectories during training.
On MNIST-like datasets (panel \textbf{b}), test log-likelihood reveals a consistent advantage for cMIM, which outperforms MIM by an average margin of 12.2\%.
This improvement was unexpected, as both methods share the same generative architecture.
We conjecture that the gain to the implicit regularization can be attributed to the contrastive term, which could reduce overfitting while maintaining generative fidelity.

\section{Related Work} \label{sec:related-work}

\paragraph{Contrastive Learning.}
Contrastive learning has become a cornerstone of self-supervised representation learning, with methods such as CPC \cite{oord2018cpc}, SimCLR \cite{chen2020simple}, and MoCo \cite{he2020momentum} demonstrating strong discriminative performance. 
These approaches typically rely on data augmentation to form positive pairs, making their success dependent on carefully chosen invariances. 
Augmentation-free contrastive methods, such as BYOL \cite{grill2020byol} and SimSiam \cite{chen2021simsiam}, avoid negatives but often require additional predictors or asymmetries for stability. 
Our work differs by integrating contrastive learning directly into a probabilistic framework, eliminating the need for augmentation or auxiliary networks.

\paragraph{Mutual Information Maximization.}
The Mutual Information Machine (MIM) \cite{livne2019mim} and follow-up works \cite{reidenbach2023molmim} maximize mutual information between inputs and latent codes while encouraging latent clustering. 
Related approaches such as Deep InfoMax \cite{hjelm2018dim} and InfoVAE \cite{zhao2017infovae} also maximize information-theoretic quantities, but typically lack a generative auto-encoding structure, or require various approximations and weighted losses which are hard to tune. 
Our method extends MIM with a contrastive component, addressing its limited discriminative power.

\paragraph{Informative Embeddings.}
Extracting hidden states from encoder–decoder models has proven effective in large language models \cite{brown2020gpt3, lee2024nv}. 
Similarly, representations from intermediate layers of auto-encoders or VAEs have been used for downstream prediction tasks \cite{alemi2018elbo}. 
We generalize this idea by introducing \emph{informative embeddings}, a systematic method to leverage decoder hidden states in probabilistic auto-encoders, demonstrating significant gains in both image and molecular tasks.

\paragraph{Unifying Generative and Discriminative Learning.}
Bridging generative modeling with discriminative performance has been a longstanding goal, explored in frameworks such as $\beta$-VAE \cite{higgins2017beta}, InfoGAN \cite{chen2016infogan}, and hybrid likelihood–contrastive models \cite{oord2018cpc}. 
Our work contributes to this line by showing that cMIM yields a single framework that maintains generative fidelity while significantly improving discriminative utility.
\section{Limitations} \label{sec:limitations}

While cMIM demonstrates clear benefits in discriminative performance and robustness to batch size, several limitations remain. 
First, we evaluate generative capacity primarily through reconstruction, leaving open the question of how cMIM performs on challenging generative tasks such as sample quality, diversity, likelihood estimation, or controlled generation. 
Second, our empirical validation is restricted to moderate-scale models and datasets; it remains to be seen how the method scales to larger architectures and high-dimensional modalities such as video or long-context language. 
Third, although cMIM removes the need for data augmentation, the choice of similarity function and temperature parameter $\tau$ may still influence results and require tuning. 
Finally, while we highlight reduced sensitivity to batch size, the method continues to benefit from larger effective numbers of negatives, which can introduce computational overhead when using memory queues or very large batches. 
These limitations motivate future work in scaling cMIM, expanding to more modalities, and further analyzing its generative behavior.
\section{Conclusions} \label{sec:conclusions}

In this paper, we introduced cMIM, a contrastive extension of the MIM framework. Unlike conventional contrastive learning, cMIM does not require positive data augmentation and exhibits reduced sensitivity to batch size compared to InfoNCE. Our experiments show that cMIM learns more discriminative features than both MIM and InfoNCE, and consistently outperforms MIM in classification and regression tasks. Moreover, cMIM maintains comparable reconstruction quality to MIM, suggesting similar performance for generative applications, though further empirical validation is needed.

We also proposed a method for extracting embeddings from encoder--decoder models, termed \textit{informative embeddings}, which improve the effectiveness of the learned representations in downstream applications.

Overall, cMIM advances the goal of unifying discriminative and generative representation learning. We hope this work provides a foundation for developing models that excel across a broad spectrum of machine learning tasks and motivates further research in this direction.

\FloatBarrier

\bibliography{paper}
\bibliographystyle{plainnat}

%%%%%%%%%%%%%%%%%%%%%%%%%%%%%%%%%%%%%%%%%%%%%%%%%%%%%%%%%%%%

\appendix

%%%%%%%%%%%%%%%%%%%%%%%%%%%%%%%%%%%%%%%%%%%%%%%%%%%%%%%%%%%

\section{Experiment Training Details} \label{sec:appendix-model-arch}

\subsection{Image Classification}

We opted for a simple architecture.
\begin{itemize}
    \item The encoder flattens the image to 784 dimensions, up-projects using a linear layer to $(784, 16)$ which is fed to a Perceiver encoder that projects it down to 400 steps $(400, 16)$. A linear layer projects the hidden dimension to 1, followed by a layer norm, and finally a linear projection from 400 to 64.
    \item The encoding distribution is a Gaussian with mean and variance predicted by linear layers from the encoder output.
    \item The decoder up-projects the 64 dimension latent code using a linear layer to $(64, 16)$ which is fed to a Perceiver encoder that projects it down to 400 steps $(400, 16)$. A linear layer projects the hidden dimension to 1, followed by a layer norm, and finally a linear projection from 400 to 784, which is reshaped back to $(28, 28)$ image dimensions.
    \item The decoding distribution is a conditional Bernoulli with logits predicted by a linear layer from the decoder output.
    \item The prior is a standard Gaussian.
\end{itemize}

All models were trained with Adam optimizer with learning rate $1e-3$ and WSD scheduler with 10\% warmup steps and 10\% decay steps, for a total of 500k steps (regardless of the batch size).

\subsection{Molecualr Property Prediction}

\textbf{Dataset:}
All models were trained using a tranche of the ZINC-15 dataset \citep{Sterling2015zinc15}, labeled as reactive and annotated, with molecular weight $\le$ 500Da and logP $\le$ 5. Of these molecules, 730M were selected at random and split into training, testing, and validation sets, with 723M molecules in the training set. We note that we do not explore the effect of model size, hyperparameters, and data on the models. Instead, we train all models on the same data using the same hyperparameters, focusing on the effect of the learning framework and the fixed-size bottleneck. For comparison, Chemformer was trained on 100M molecules from ZINC-15 \citep{Sterling2015zinc15} -- 20X the size of the dataset used to train CDDD (72M from ZINC-15 and PubChem \citep{Kim2018pubchem}). MoLFormer-XL was trained on 1.1 billion molecules from the PubChem and ZINC datasets.

\textbf{Data augmentation:}
Following \citet{Irwin_2022}, we used two augmentation methods: masking, and SMILES enumeration \citep{Weininger1988smiles}. Masking is as described for the BART MLM denoising objective, with 10\% of the tokens being masked, and was only used during the training of MegaMolBART. 
In addition, MegaMolBART, PerBART, and MolVAE used SMILES enumeration where the encoder and decoder received different valid permutations of the input SMILES string.
MolMIM was the only model to see an increase in performance when both the encoder and decoder received the same input SMILES permutation,
simplifying the training procedure.

\textbf{Model details:}
We implemented all models with NeMo Megatron toolkit \citep{Kuchaiev2019nemo}.
We used a RegEx tokenizer with 523 tokens \citep{Bird2009nltk}.
All models had 6 layers in the encoder and 6 layers in the decoder, with a hidden size of 512, 8 attention heads, and a feed-forward dimension of 2048.
The Perceiver-based models also required defining K, the hidden length, which relates to the hidden dimension by $H = K \times D$ where $H$ is the total hidden dimension, and $D$ is the model dimension (Fig. \ref{fig:informative-embeddings}).
MegaMolBART had $58.9 M$ parameters, PerBART had $64.6 M$, and MolVAE and MolMIM had $65.2 M$.
We used greedy decoding in all experiments.
We note that we trained MolVAE using the loss of $\beta$-VAE \citep{higgins2017beta} where we scaled the KL divergence term with $\beta = \frac{1}{D}$ where $D$ is the hidden dimensions.

\textbf{Optimization:}
We use ADAM optimizer \citep{Kingma2015adam} with a learning rate of 1.0, betas of 0.9 and 0.999, weight decay of 0.0, and an epsilon value of 1.0e-8.
We used Noam learning rate scheduler \citep{Vaswani2017attention} with a warm-up ratio of 0.008, and a minimum learning rate of 1e-5.
During training, we used a maximum sequence length of 512, dropout of 0.1, local batch size of 256, and global batch size of 16384.
All models were trained for 1,000,000 steps with fp16 precision for 40 hours on 4 nodes with 16 GPU/node (Tesla V100 32GB).
MolVAE was trained using $\beta$-VAE \citep{higgins2017beta} with $\beta = \frac{1}{H}$ where $H$ is the number of hidden dimensions.
We have found this choice to provide a reasonable balance between the rate and distortion (see \citet{alemi2018elbo} for details). It is important to note that MolMIM does not require the same $\beta$ hyperparameter tuning as done for VAE.

\section{Additional Results} \label{sec:appendix-additional-results}

\subsection{MNIST-like Image Classification}

\begin{figure}[ht]
    \centering
    \begin{subfigure}[b]{0.45\textwidth}
        \centering
        \includegraphics[width=\textwidth]{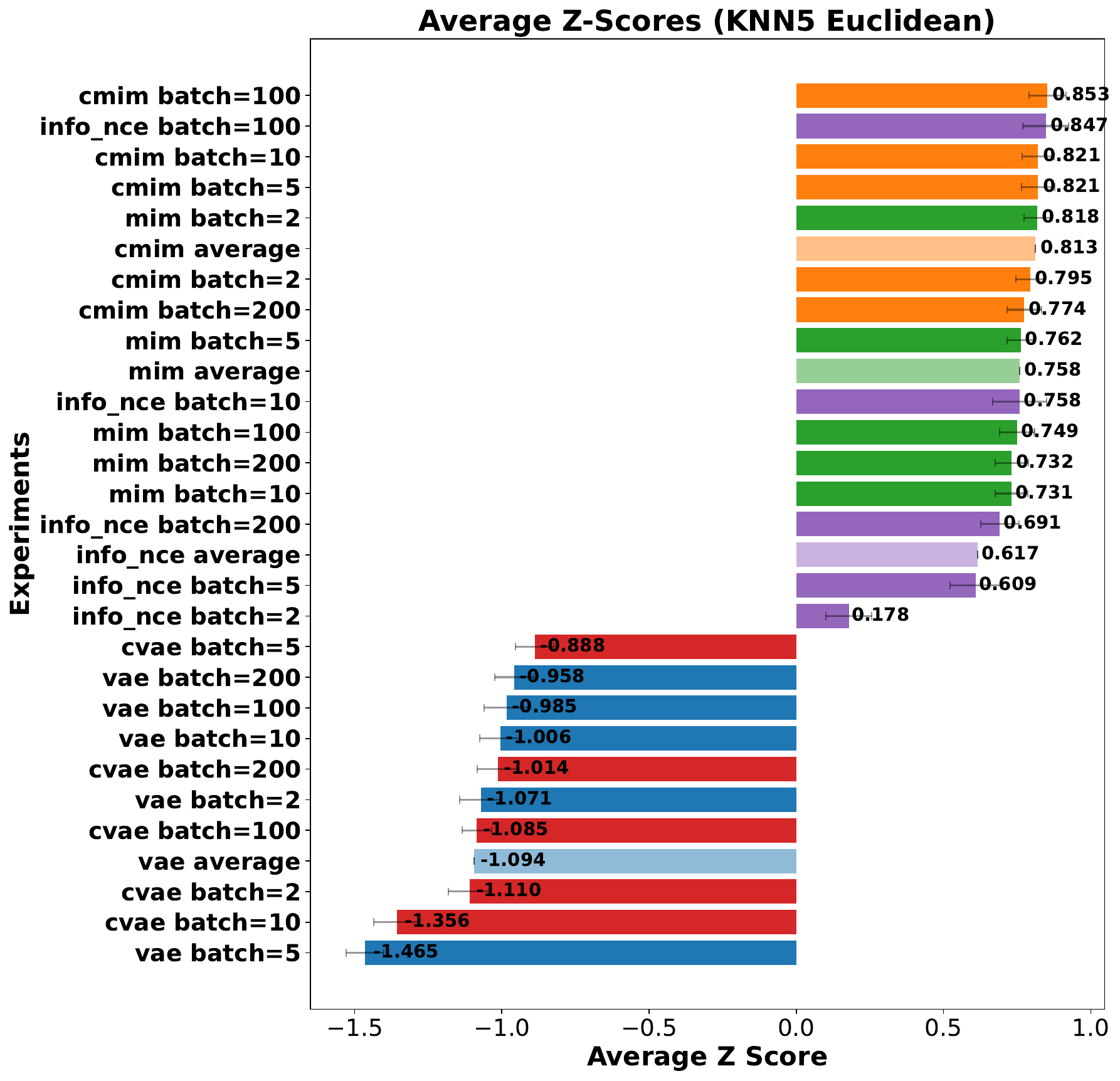}
        \caption{KNN5 Euclidean}
        \label{fig:mnist-z-scores-knn5-euclidean}
    \end{subfigure}
    \hfill
    \begin{subfigure}[b]{0.45\textwidth}
        \centering
        \includegraphics[width=\textwidth]{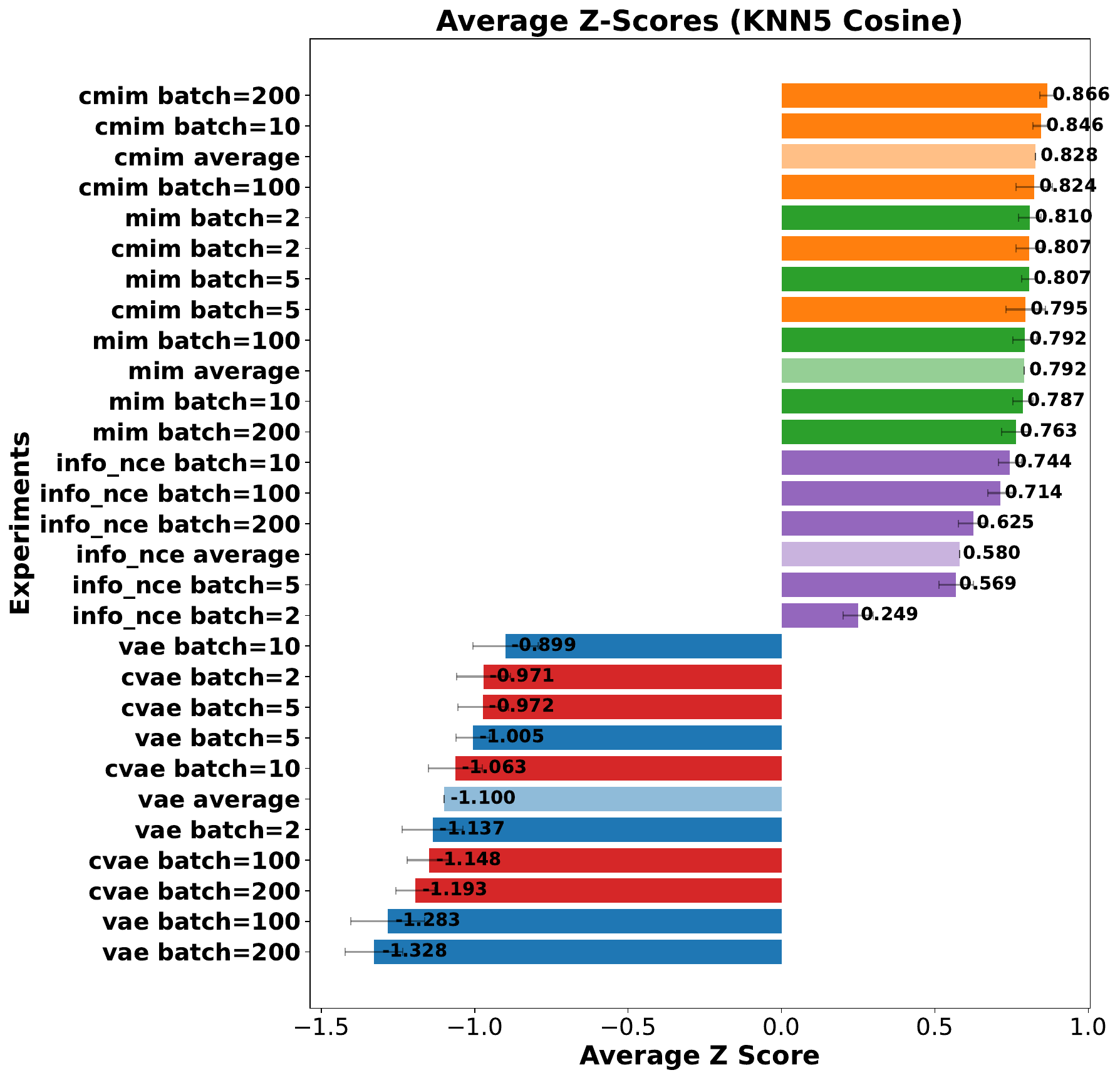}
        \caption{KNN5 Cosine}
        \label{fig:mnist-z-scores-knn5-cosine}
    \end{subfigure}
    
    \vspace{0.5cm}
    
    \begin{subfigure}[b]{0.45\textwidth}
        \centering
        \includegraphics[width=\textwidth]{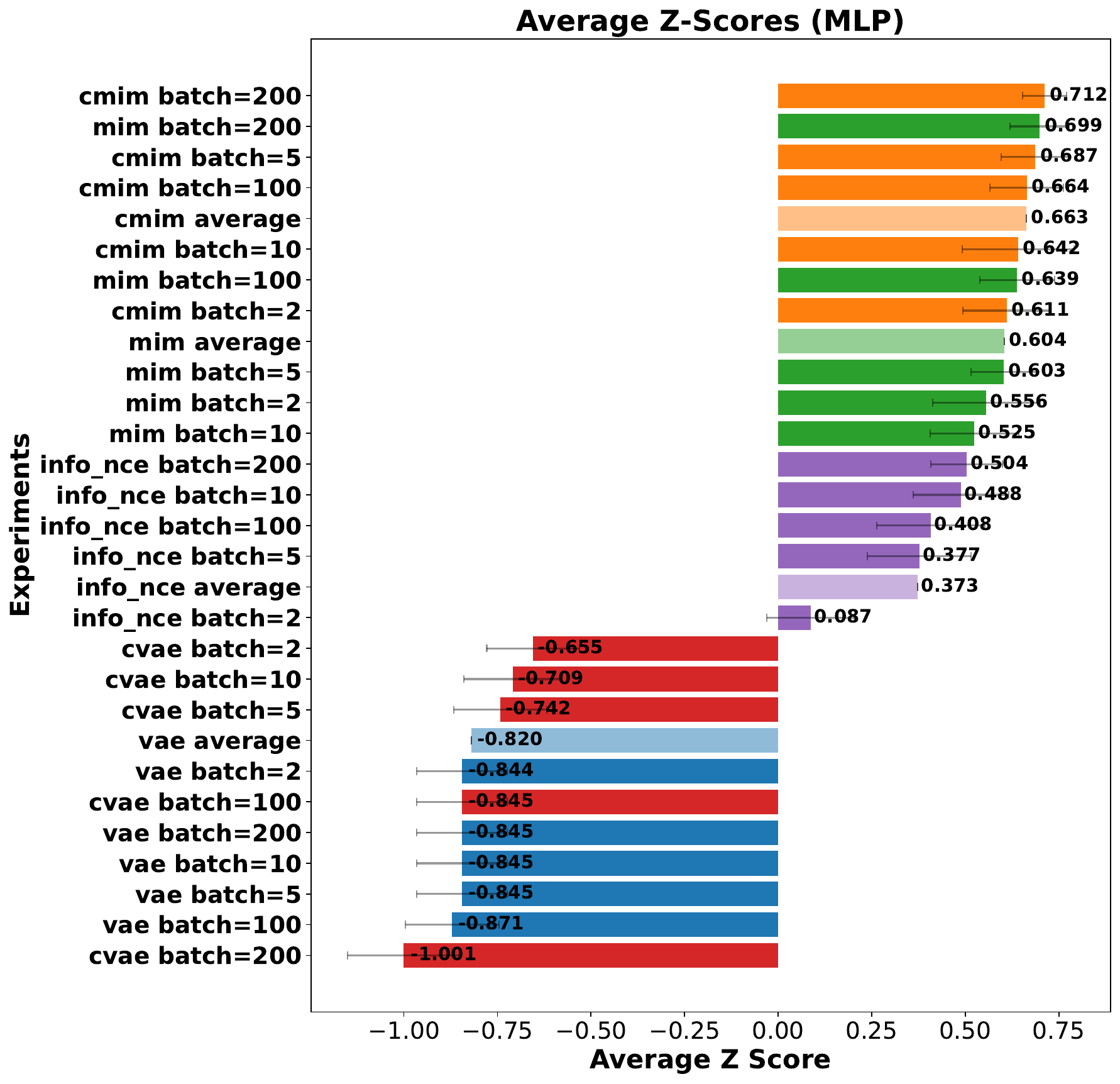}
        \caption{MLP}
        \label{fig:mnist-z-scores-mlp}
    \end{subfigure}
    \caption{Z-scores with error bars for MNIST-like image classification tasks using different evaluation methods.}
    \label{fig:mnist-z-scores}
\end{figure}

\begin{figure}[ht]
    \centering
    \begin{subfigure}[b]{0.45\textwidth}
        \centering
        \includegraphics[width=\textwidth]{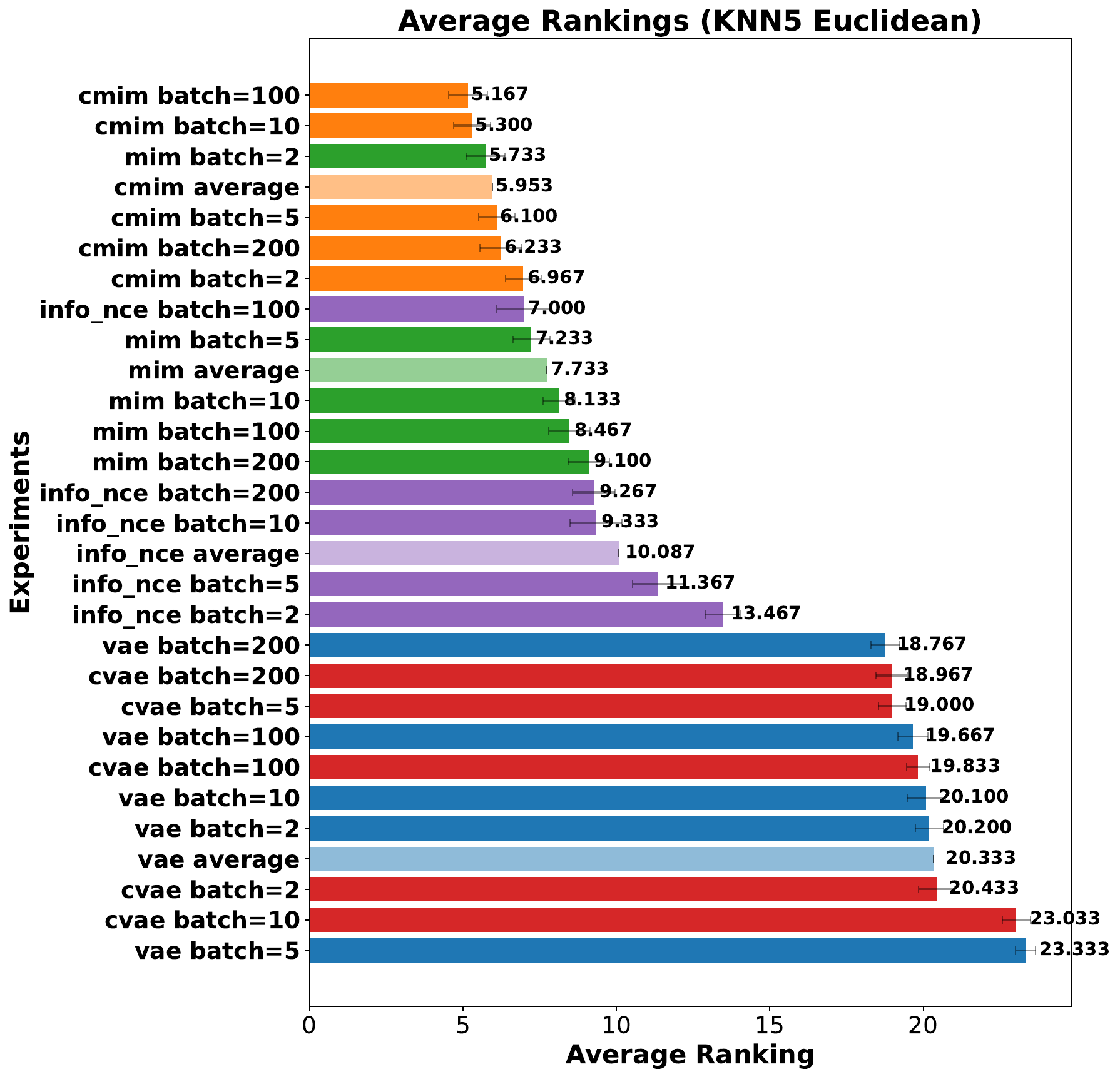}
        \caption{KNN5 Euclidean}
        \label{fig:mnist-rankings-knn5-euclidean}
    \end{subfigure}
    \hfill
    \begin{subfigure}[b]{0.45\textwidth}
        \centering
        \includegraphics[width=\textwidth]{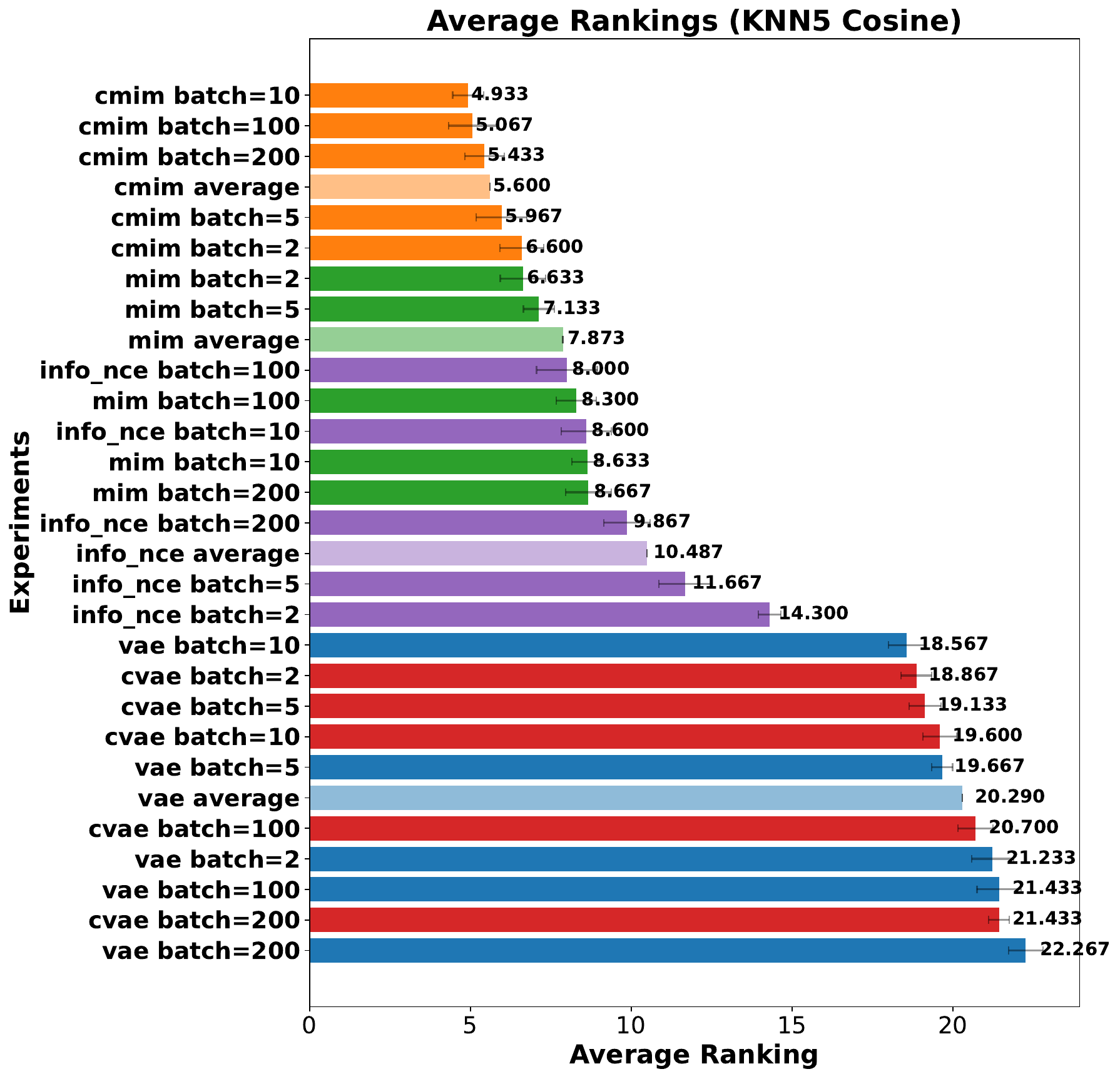}
        \caption{KNN5 Cosine}
        \label{fig:mnist-rankings-knn5-cosine}
    \end{subfigure}
    
    \vspace{0.5cm}
    
    \begin{subfigure}[b]{0.45\textwidth}
        \centering
        \includegraphics[width=\textwidth]{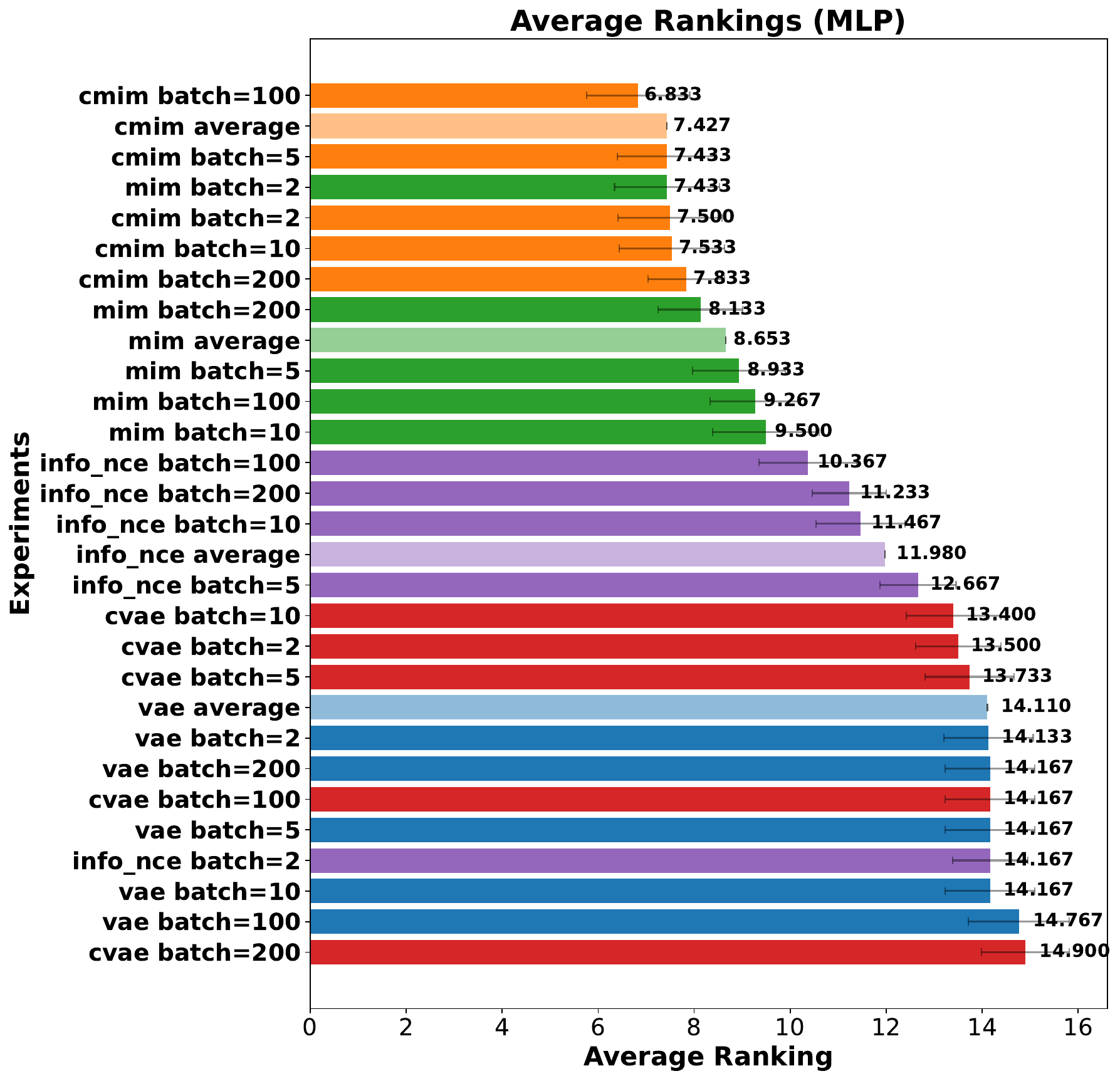}
        \caption{MLP}
        \label{fig:mnist-rankings-mlp}
    \end{subfigure}
    \caption{Rankings with error bars for MNIST-like image classification tasks using different evaluation methods.}
    \label{fig:mnist-rankings}
\end{figure}

% Optionally include supplemental material (complete proofs, additional experiments and plots) in appendix.
% All such materials \textbf{SHOULD be included in the main submission.}

%%%%%%%%%%%%%%%%%%%%%%%%%%%%%%%%%%%%%%%%%%%%%%%%%%%%%%%%%%%%

\end{document}